\pdfoutput=1
\PassOptionsToPackage{dvipsnames}{xcolor}  

\documentclass[11pt]{article}
\usepackage[final]{acl}  

\usepackage[T1]{fontenc}
\usepackage[utf8]{inputenc}
\usepackage{microtype}
\usepackage{times}
\usepackage{latexsym}
\usepackage{inconsolata}

\usepackage{amsmath}
\usepackage{amssymb}
\usepackage{amsfonts}

\usepackage{graphicx}
\usepackage{tikz}
\usepackage{tcolorbox}
\usepackage{pifont}

\usepackage{booktabs}
\usepackage{multirow}
\usepackage{makecell}
\usepackage{array}

\usepackage{algorithmic}

\usepackage{enumitem}

\newcommand{\blackcircled}[1]{%
  \tikz[baseline=(char.base)]{
    \node[shape=circle,draw=black,fill=black,inner sep=0.3pt] 
      (char) {\textcolor{white}{\sffamily\bfseries #1}};}}

\title{CoopetitiveV: Leveraging LLM-powered Coopetitive Multi-Agent Prompting for High-quality Verilog Generation}

\author{
  Zhendong Mi\textsuperscript{1}, 
  Renming Zheng\textsuperscript{1}, 
  Haowen Zhong\textsuperscript{2}, 
  Yue Sun\textsuperscript{3}, \\
  \textbf{Seth Kneeland}\textsuperscript{4}, 
  \textbf{Sayan Moitra}\textsuperscript{4},
  \textbf{Ken Kutzer}\textsuperscript{4}, 
  \textbf{Zhaozhuo Xu}\textsuperscript{1}, 
  \textbf{Shaoyi Huang}\textsuperscript{1} \\
  \textsuperscript{1}Stevens Institute of Technology \quad
  \textsuperscript{2}University of Washington \quad \\
  \textsuperscript{3}Amazon \quad
  \textsuperscript{4}SambaNova Systems\\
  \texttt{\small \{zmi2, rzheng3, zxu79, shuang59\}@stevens.edu}, 
  \texttt{\small haowenz@uw.edu}, \\
   \texttt{\small \{seth.kneeland, sayan.moitra, ken.kutzer\}@sambanovasystems.com}
}

\begin{document}
\maketitle
\begin{abstract}
Recent advances in agentic LLMs have demonstrated great capabilities in Verilog code generation. However, existing approaches either use LLM-assisted single-agent prompting  or cooperation-only multi-agent learning, which will lead to: (i) \textit{Degeneration issue for single-agent learning}: characterized by diminished error detection and correction capabilities; (ii) \textit{Error propagation in cooperation-only multi-agent learning}: erroneous information from the former agent will be propagated to the latter through prompts, which can make the latter agents generate buggy code.
In this paper, we propose an LLM-based coopetitive multi-agent prompting framework,
in which the agents cannot collaborate with each other to form the generation pipeline, but also create a healthy competitive mechanism to improve the generating quality.
Our experimental results show that the coopetitive multi-agent framework can effectively mitigate the degeneration risk and reduce the error propagation while improving code error correction capabilities, resulting in higher quality Verilog code generation. The effectiveness of our approach is validated through extensive experiments. On VerilogEval Machine and Human dataset, CoopetitiveV+GPT-4 achieves 99.2\% and 99.1\% pass@10 scores, respectively. While on RTLLM,  CoopetitiveV+GPT-4 obtains 100\% syntax and 99.9\% functionality pass@5 scores.
\end{abstract}

\section{Introduction}

\begin{figure}
    \centering
    \includegraphics[width=0.99\linewidth]{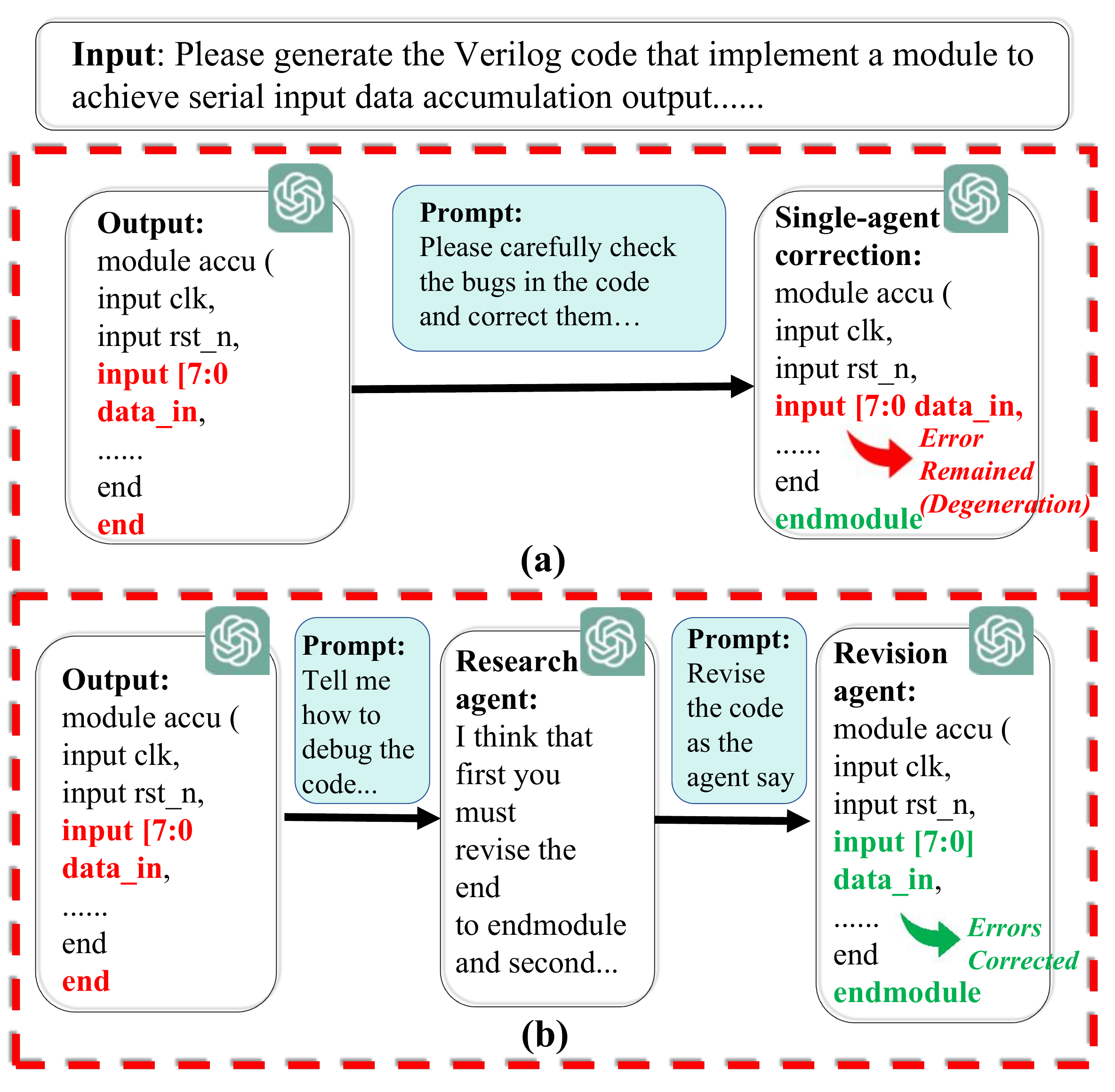}
    \caption{(a) Degeneration issue in single-agent based Verilog code generation; (b) Multi-agent mechanism can mitigate the degeneration issue}
    \label{fig:1}
\end{figure}

As semiconductor technology advances to smaller process nodes (7nm, 5nm, 3nm, and beyond), electronic design automation (EDA) faces increasing challenges due to the escalating design complexity, increasing human resource constraints, and intensifying time-to-market pressure. Hardware description language (HDL) code generation, as a fundamental EDA task, particularly exemplifies these challenges. In recent years, large language models (LLMs) have captured extensive attention due to their significant performance across various tasks~\cite{dubey2024llama3herdmodels, openai2024gpt4technicalreport, touvron2023llama2openfoundation} and have emerged as a promising solution, demonstrating significant potential in automating various EDA tasks, especially in HDL code generation where traditional approaches struggle to scale~\cite{liu2024chipnemodomainadaptedllmschip}.

\begin{figure*}[!ht]
    \centering
    \includegraphics[width=0.99\textwidth]{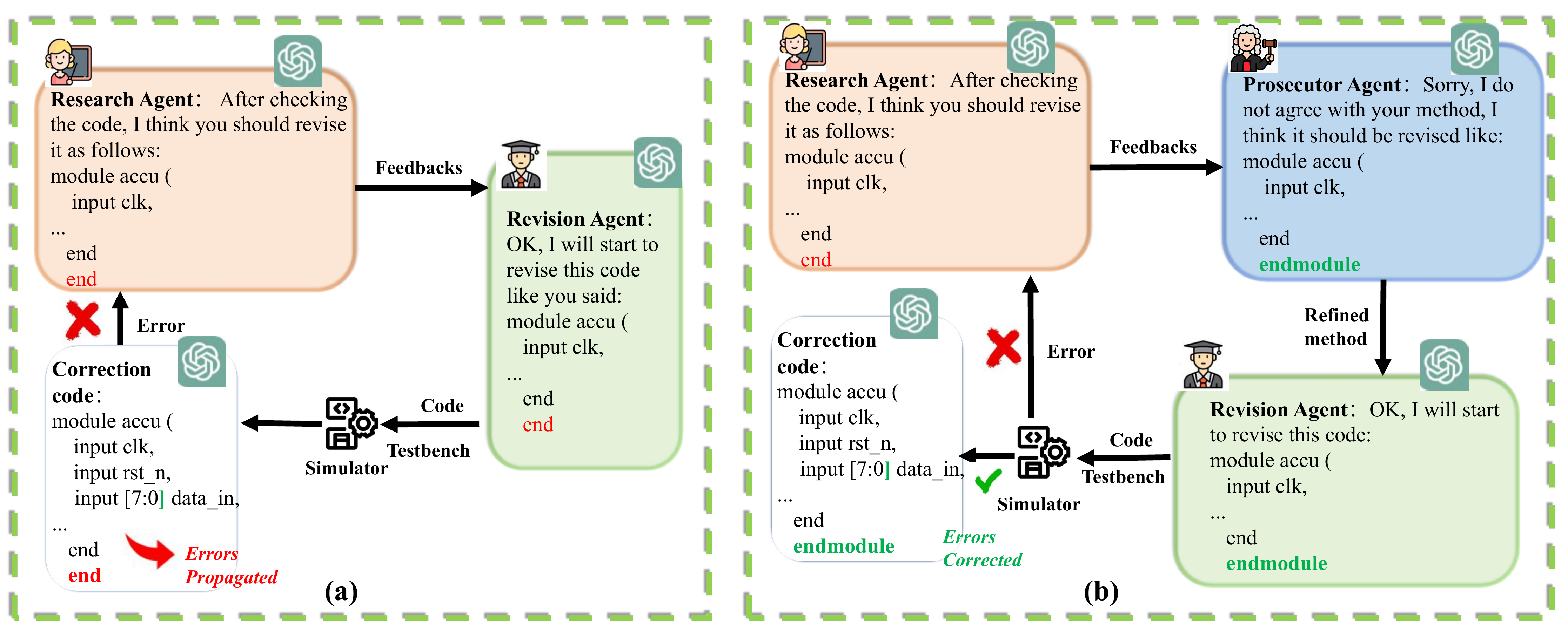}
    \caption{LLM-based multi-agent learning for Verilog generation with (a) Cooperation-only mechanism; (b) Coopetitive mechanism}
    \vspace{-0.1in}
    \label{fig:cooperation-only}
\end{figure*}

Prompting~\cite{olausson2023self, Zhang2024LearningTC, Huang2024TowardsLV} has been adopted for
HDL generation and error correction 
as the crafted prompts can guide the pretrained LLM in generating desired outputs with a few examples without modifying the model. While effective, existing approaches typically
employ LLM-assisted single-agent prompting techniques~\cite{olausson2023self, Zhang2024LearningTC, Huang2024TowardsLV}, where the model selectively performs code generation, self-execution, and self-correction in sequence. However, we observe the single-agent prompting has a degeneration issue, which is characterized by deteriorating generative performance and diminished error detection and correction capabilities. As shown in Figure~\ref{fig:1}(a), the error remains in the code when using LLM-assisted single-agent for correction. Although LLM-based multi-agent learning can potentially address the issue as shown in Figure~\ref{fig:1}(b), we observe the error propagation issue in the cooperative-only-based multi-agent learning pipeline, in which the error information generated by the former agents can be sent to the latter agents via prompts, thus harming the final generation effects, as shown in Figure \ref{fig:cooperation-only}(a).

In this work, we introduce \textbf{CoopetitiveV}, which leverages LLM-powered \underline{coopetitive} multi-agent prompting for high-quality \underline{V}erilog generation. Instead of utilizing single-agent for both code generation and error correction, CoopetitiveV utilizes a multi-agent architecture (multiple LLM agents are employed for different tasks, e.g., code generation, testbench generation, error correction strategies generation, code correction, and testbench correction) to reduce the degeneration issue. 
Moreover, we integrate the cooperation-competition mechanism into the framework: a research agent that identifies and analyzes the errors while providing correction strategies for both code and testbench, and a prosecutor agent to judge the contributions of the research agents while generating the refined strategies for correction. Then, the dual revision agents will apply the prosecutor's refined suggestions to correct the errors in the code and testbench, respectively.
This competitive mechanism can help to
mitigate the cascading failures that may arise in purely cooperation-only systems where a single misstep by one agent leads to compounding errors throughout the whole correction process \cite{tran2025multi, liang2023encouraging, chen2024llmarena}. Figure \ref{fig:cooperation-only} shows the comparison between
the cooperation-only mechanism (Figure \ref{fig:cooperation-only}(a)) and the coopetitive mechanism (Figure \ref{fig:cooperation-only}(b)). 
We summarize our contributions as follows:

\begin{figure*}[!ht]
    \centering
    \includegraphics[width=1.00\textwidth]{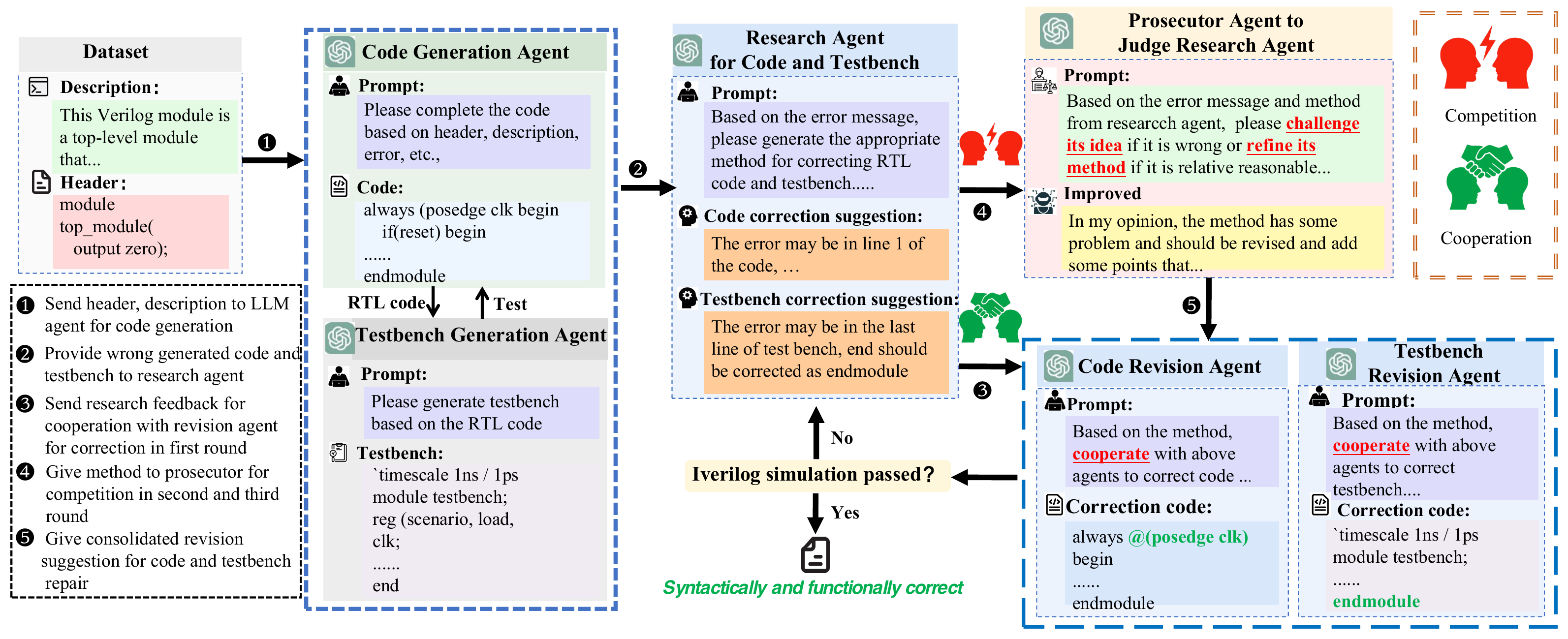}
    \caption{The proposed LLM-powered multi-agent prompting framework for Verilog generation}
    \label{fig:main}
\end{figure*}


\begin{itemize}[leftmargin=*]
\setlength{\itemsep}{0.05in}
    \item We observe the degeneration issue in LLM-assisted 
          single-agent prompting for Verilog generation, 
          which is characterized by deteriorating generative performance 
          and diminished error detection and correction capabilities.
    \item We observe the error propagation in LLM-assisted cooperation-only multi-agent learning pipeline in Verilog code correction, in which the latter LLM agents will receive the wrong messages from the former agents and make corresponding suboptimal correction of codes.
    \item We propose CoopetitiveV, an LLM-powered interactive multi-agent prompting framework. Within the framework, we propose the cooperation and competition mechanism which can effectively help to address both the degeneration issue and the error propagation issue, thus leading to improved Verilog generation quality.
    \item We conduct extensive experiments across various datasets, including VerilogEval-Machine, VerilogEval-Human, Verilog-V2, and RTLLM to demonstrate the effectiveness of CoopetitiveV in Verilog generation. We show that CoopetitiveV+GPT-4 can achieve 93.9\% and 89.3\% pass@1 scores on Verilog-Machine and Verilog-Human datasets, 
with up to 40\%-50\% higher accuracy compared with the baselines.

    
\end{itemize}

\begin{figure}
    \centering
    \includegraphics[width=0.99\linewidth]{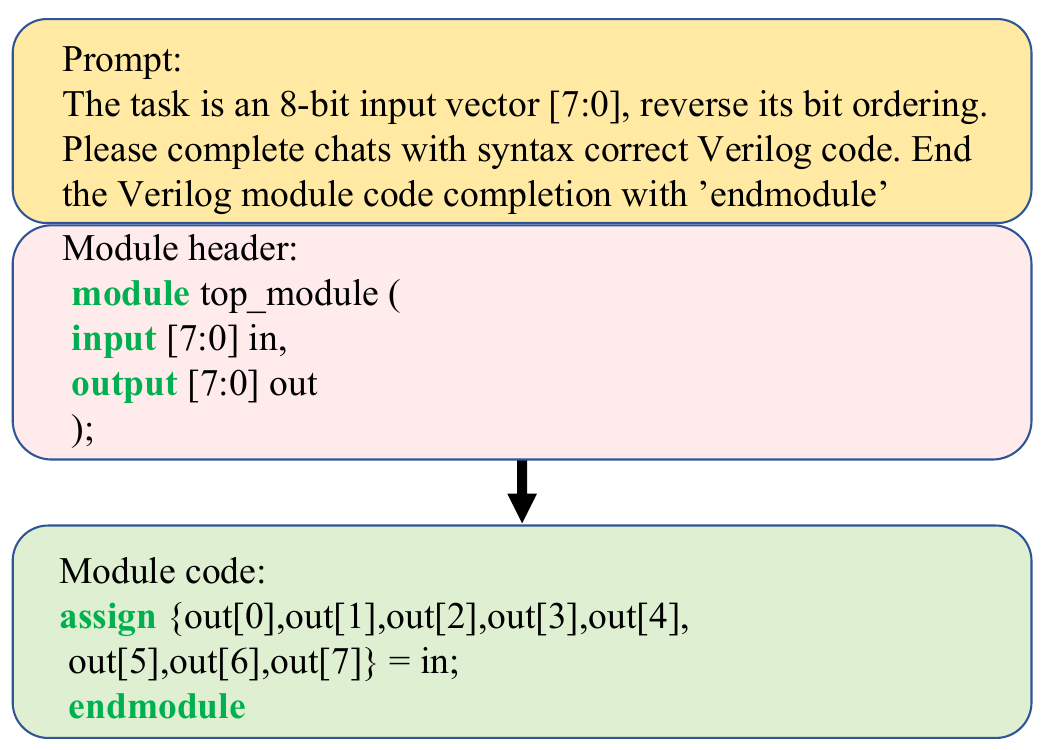}
    \caption{Verilog generation task}
    \label{fig:task}
\end{figure}

\section{Method}

\subsection{Task Description}
While LLM research has made considerable progress in software programming languages such as Python and JavaScript, its application to hardware description languages like Verilog remains relatively underexplored. 
In this work,
our task is to obtain syntactically and functionally accurate Verilog code based on provided problem description and module headers as shown in Figure~\ref{fig:task}. 
In general, our pipeline consists of three sub-tasks: 1) Verilog code and testbench generation;
2) code and testbench error correction strategies generation;
3) code and testbench error correction.

\subsection{Overview of the Proposed LLM-based Multi-agent Verilog Generation Pipeline}


We present an LLM-based multi-agent prompting pipeline that facilitates a collaborative and competitive interactive workflow for code generation, verification, and correction, as shown in Figure~\ref{fig:main}. The system architecture integrates multiple specialized agents with Icarus Verilog (Iverilog), an open-source hardware description language simulation and synthesis tool.
Multiple specialized agents in the framework include a code generation agent, a testbench generation agent, a research agent, a prosecutor agent, and dual revision agents.
More specifically, our pipeline consists of the following steps:
\blackcircled{1} \textbf{\textit{Verilog module code generation and testbench generation}}: 
with the inputs of description and header, the code generation agent generates Verilog code that adheres to the specified interface requirements and functional constraints. Subsequently, we employ AutoBench~\cite{qiu2024autobench} for testbench generation and functionality verification. 
\blackcircled{2} \textbf{\textit{Research agent providing error correction suggestions for both code and testbench}}: 
if the code cannot pass Autobench simulation, the research agent automatically receives the wrong code and testbench and then analyzes the existing errors including syntax errors, logical inconsistencies, testbench verification issues, etc. Then, it generates specific correction suggestions to address the issues.
\blackcircled{3} \textbf{\textit{Cooperation-only Verilog code and testbench error correction}}: 
at first round of error correction,
the revision agents will receive revision suggestions 
from the research agent directly, establishing a cooperation-only mechanism for correcting erroneous code together with the research agent.
\blackcircled{4} \textbf{\textit{Prosecutor agent challenging the research agent}}: if the revised code still fails to pass the simulation, a prosecutor agent will judge the effectiveness of the strategy proposed by the research agent and generate an improved strategy, fostering a healthy competition between the two agents and this will form a competitive mechanism between multi-agents.
\blackcircled{5} \textbf{\textit{Revision agents correct code and testbench}}: the improved strategy will be sent to the revision agent for code correction and refinement.






\subsection{Multi-agent Coopetitive Verilog Code Generation}
\subsubsection{Problem Setting}

Our LLM-based multi-agent learning pipeline is designed to generate Verilog code and corresponding testbench as correctly as possible through an \textit{error-and-trial} process.
The pipeline has two elements in the inputs: the prompt base $\textbf{\textit{p}}_0$ and the Verilog module information $\textbf{\textit{m}}$. The latter encompasses both the comprehensive module description and its corresponding header specification. These components serve as essential prerequisites for the subsequent stages of the generation process.
Leveraging the combination of 
$\textbf{\textit{p}}_0$ and $\textbf{\textit{m}}$ 
as prompt $\textit{\textbf{p}}$, the code generation agent \textbf{\textit{C}} generates Verilog module implementations $\textbf{\textit{Code}}_0$ that adheres to the specified interface requirements and functional constraints, which can be formulated as:
\begin{align*}
    \textit{\textbf{p}} = \textit{\textbf{p}}_0 + \textbf{\textit{m}}, \quad
    \textbf{\textit{Code}}_0 = \textbf{\textit{C}}(\textbf{\textit{p}})
\end{align*}
   



Subsequently, we employ AutoBench~\cite{qiu2024autobench} for testbench generation and comprehensive function verification. We use $\textbf{\textit{TB}}_0$ as the generated testbench. Specifically, 
the AutoBench, denoted as \textit{\textbf{A}}, not only synthesizes appropriate verification environments by generating the testbench but also performs code compilation through the embedded Iverilog, providing compilation results for both the Verilog code and its corresponding testbench. Thus we can obtain the testbench generation $\textbf{\textit{TB}}_0$, $\textbf{\textit{Pass}}_0$ results and the detailed code and testbench error information $\textbf{\textit{E}}_0$ if $\textbf{\textit{Pass}}_0=\textbf{\textit{False}}$ through the entire process of Autobench. We formulate the Autobench generation as well as the code and testbench syntax verification process as follows:

\begin{align*}          \textbf{\textit{TB}}_0,\textbf{\textit{Pass}}_0,\textbf{\textit{E}}_0 = \textbf{\textit{A}}(\textbf{\textit{Code}}_0,\textbf{\textit{m}})
\end{align*}


We consider the process above as 
the settings or prerequisite of our Verilog code and testbench generation and correction pipeline.
%

\subsubsection{From Single-agent to Multi-agent}

\begin{tcolorbox}[
 colback=green!5,
 colframe=green!20,
 boxrule=0.5pt,
 arc=4pt,
 boxsep=5pt,
 left=6pt, right=6pt,
 top=6pt, bottom=6pt,
 fonttitle=\bfseries,
 coltitle=black,
 colbacktitle=green!10,
 colframe=green!20,
 title=Findings 1
]
\textit{We observe \textbf{degeneration issue} in the single-agent framework for code correction, where the agent will repeat the previous wrong responses, which will be highly inefficient to generate and debug accurate Verilog code.
}
\end{tcolorbox}

When employing a single-agent framework for Verilog code generation and refinement, we observed that the agent would occasionally repeat its previous responses, retain erroneous segments, and fail to identify or correct issues within the code. Consequently, we shifted to a multi-agent paradigm, leveraging collaborative mechanisms to guide the language model in detecting and rectifying errors more effectively.

\subsubsection{Error Propagation Exists in Cooperation-only Multi-agent Learning}

\begin{tcolorbox}[
 colback=green!5,
 colframe=green!20,
 boxrule=0.5pt,
 arc=4pt,
 boxsep=5pt,
 left=6pt, right=6pt,
 top=6pt, bottom=6pt,
 fonttitle=\bfseries,
 coltitle=black,
 colbacktitle=green!10,
 colframe=green!20,
 title=Findings 2
]
\textit{We observe \textbf{error propagation issue} in the cooperation-only multi-agent mechanism, where the Research Agent will generate code error correction strategy and provide it to the Code Revision Agent. However, errors in the generated strategy from the former (Research Agent) will lead to errors in the output of the latter (Code Revision Agent).
}
\end{tcolorbox}


When employing the cooperation-only multi-agent framework, as depicted in Figure \ref{fig:cooperation-only}(a), for Verilog code generation and refinement, we observed that if the research agent introduces an error or proposes a solution that lacks completeness during the initial suggestion phase, this flaw will be propagated to the following revision agents. The revision agents may then incorrectly assume that the research agent’s suggestion is correct and comprehensive. As a consequence, during the debugging phase, the revision agents can apply flawed or incomplete modifications to the code and testbench, ultimately leading to suboptimal outcomes or even a complete breakdown of the code revision process.

\subsubsection{Coopetitive Mechanism }

In this section, we aim to answer the question \textit{how to solve the error propagation in cooperation-only multi-agent mechanism for improved Verilog code generation?}
Inspired by debate competition between humans, we introduce a competitive mechanism among multiple  agents. This mechanism encourages agents to identify each other's mistakes, continuously refine and complement proposed solutions, and ultimately achieve promising results in Verilog code generation and correction tasks.

\subsubsection{The Cooperative Module}
In this module, the research agent will cooperate directly with two revision agents to modify the Verilog code and its corresponding testbench when it is the first round of correction. The research agent will provide explicit code modification suggestions to the revision agents, enabling them to implement the changes directly.



\noindent\textbf{\textit{Research agent:}}
We introduce a specialized research agent $\textit{\textbf{T}}$ that functions as an expert for Verilog code error analysis and correction strategy development. 
When syntax or functionality validation fails 
for either the Verilog code or its associated testbench, $\textit{\textbf{T}}$ automatically processes the error information through a systematic analysis
of both the error code and testbench and the corresponding error messages, culminating in the development of a correction strategy $\textit{\textbf{S}}$ 
that provides targeted recommendations for code and testbench error correction.
The research agent is designed with multiple specialized functions, including
syntax error examination, logical inconsistency detection, testbench verification 
assessment, 
etc. Through these capabilities, 
$\textit{\textbf{T}}$ develops target-oriented error
correction strategies that address identified as well as
potential bugs 
while ensuring compliance with hardware description language specifications and verification requirements.
The error correction process follows an iterative approach, where $k$ represents the iteration counter with a predetermined upper limit (We set 2 as the upper limit of \textbf{\textit{k}} in our experiment). In the base case of $\textbf{\textit{k}} = 0$, $\textit{\textbf{T}}$ generates the initial correction strategy by analyzing the generated code $\textbf{\textit{Code}}_0$,  testbench $\textbf{\textit{TB}}_0$, and the error information $\textbf{\textit{E}}_0$.
%
%
In the case of $k > 0$, we 
utilize the combined updated Verilog code and testbench from previous round along with associated error information as inputs of the research agent.
We denote them as $\textbf{\textit{Code}}_k$, $\textbf{\textit{TB}}_k$ and $\textbf{\textit{E}}_k$, respectively. 
The error correction strategy generation process is
$\textbf{\textit{S}} = (\textbf{\textit{S}}_c^{k}, \textbf{\textit{S}}_t^{k})$.
\begin{align*}
   \textbf{\textit{S}} = \textbf{\textit{T}}(\textbf{\textit{p}}, \textbf{\textit{Code}}_k, \textbf{\textit{TB}}_k, \textbf{\textit{E}}_k), k=0, 1, 2
\end{align*}


\noindent\textbf{\textit{Verilog code revision agent and testbench revision agent:}}
The error correction phase is executed through a dual-agent parallel processing
framework. Two specialized revision agents operate concurrently to execute the correction strategies formulated by the research agent or prosecutor agent.
In the $\textit{k}-th$ iteration, the correction strategy $(\textbf{\textit{S}}_c^k,\textbf{\textit{S}}_t^k)$ from the research agent or $(\textbf{\textit{C}}_c^k,\textbf{\textit{C}}_t^k)$ from the prosecutor agent is
distributed to two dedicated agents:
a) \textit{Code revision agent} $\textbf{\textit{L}}_c$ specializing in
Verilog module code correction;
b) \textit{Testbench revision agent} $\textbf{\textit{L}}_t$ dedicated to testbench refinement. 
When $\textbf{\textit{Pass}}_k=\textbf{\textit{False}}$ in the previous simulation step, the dual learner agents generate the corrected Verilog code $\textbf{\textit{Code}}_{k+1}$ and the refined testbench $\textbf{\textit{TB}}_{k+1}$. However, in this cooperative module, specifically when $\textbf{\textit{k}} = 0$, the two revision agents work in direct coordination with the research agent, receiving strategies $(\textbf{\textit{S}}_c^k,\textbf{\textit{S}}_t^k)$ directly from the research agent to guide their modifications:
\begin{align*}
   \textbf{\textit{Code}}_{k+1} = \textbf{\textit{L}}_c(\textbf{\textit{p}}, \textbf{\textit{Code}}_k, \textbf{\textit{TB}}_k, \textbf{\textit{E}}_k, \textbf{\textit{S}}_c^{k}), k = 0 \\
   \textbf{\textit{TB}}_{k+1} = \textbf{\textit{L}}_t(\textbf{\textit{p}}, \textbf{\textit{Code}}_k, \textbf{\textit{TB}}_k, \textbf{\textit{E}}_k, \textbf{\textit{S}}_t^{k}), k = 0
\end{align*}


The revision agents implement the error correction in parallel within their designated domains based on the strategies from the research agent, offering several key benefits:
1) \textit{Parallel processing efficiency:} The concurrent operation of agents enables simultaneous refinement of module code and testbench components, optimizing the overall error correction timeline;
2) \textit{Domain isolation:} The strict separation between $\textbf{\textit{L}}_c$ and $\textbf{\textit{L}}_t$ agents ensures that modifications to module implementation and verification logic proceed without cross-interference or unintended dependencies.
3) \textit{Process integrity:} Each agent maintains dedicated control over its respective component, enabling focused error resolution while preserving the structural integrity of both code and testbench elements.
Overall, the systematic approach to parallel error correction enhances both the efficiency and reliability of the error correction process.

\subsubsection{The Competitive Module}

\noindent\textbf{\textit{Prosecutor agent:}} 
 The prosecutor agent is positioned between the research agent and the revision agents. Activation of the prosecutor agent occurs conditionally after the first round of code correction ($k > 0$), when the code fails to pass Iverilog tests in the initial correction attempt. Upon activation, the prosecutor agent evaluates strategies
$\textbf{\textit{S}} = (\textbf{\textit{S}}_c^k,\textbf{\textit{S}}_t^k)$  proposed by the research agent through a systematic scoring mechanism. This evaluation process involves identifying deficiencies in lower-scored strategies and proposing corresponding improvements. In cases where the research agent's strategy is deemed sound, the prosecutor agent provides supplementary insights and generates a comprehensive summary of the proposed approach. The mechanism can be formulated as follows:
\begin{align*}
   \textbf{\textit{S}} = (\textbf{\textit{S}}_c^{k}, \textbf{\textit{S}}_t^{k}), \quad
   \textbf{\textit{C}} = (\textbf{\textit{C}}_c^{k}, \textbf{\textit{C}}_t^{k})\\
   \textbf{\textit{C}} = \textbf{\textit{P}}(\textbf{\textit{p}}, \textbf{\textit{Code}}_k, \textbf{\textit{TB}}_k, \textbf{\textit{E}}_k, \textbf{\textit{S}}), k= 1, 2
\end{align*}
Where \textbf{\textit{C}} denotes the final generated strategies which are the combination of more comprehensive solutions and suggestions for correction.
After the prosecutor agent provides more comprehensive suggestions, the two revision agents will receive these suggestions \textbf{\textit{C}} in the subsequent rounds where \textbf{\textit{k}} > 0, and accordingly apply the modifications to the Verilog code and the testbench:
\begin{align*}
   \textbf{\textit{Code}}_{k+1} = \textbf{\textit{L}}_c(\textbf{\textit{p}}, \textbf{\textit{Code}}_k, \textbf{\textit{TB}}_k, \textbf{\textit{E}}_k, \textbf{\textit{C}}_c^{k}), k = 1,2 \\
   \textbf{\textit{TB}}_{k+1} = \textbf{\textit{L}}_t(\textbf{\textit{p}}, \textbf{\textit{Code}}_k, \textbf{\textit{TB}}_k, \textbf{\textit{E}}_k, \textbf{\textit{C}}_t^{k}), k = 1,2
\end{align*}

\subsection{Interactive Multi-agent Learning for Verilog Generation}

Following 
the multi-agents setup, we implement an
interactive multi-agent learning pipeline for Verilog generation and correction. 
The Verilog code and testbench will first be generated, along with corresponding simulation information. 
Subsequently, the inner cycle of Verilog code and testbench error correction will be carried out with information of $\textit{\textbf{p}}, \textbf{\textit{m}}, \textbf{\textit{Code}}_0, \textbf{\textit{TB}}_0, \textbf{\textit{E}}_k, \textbf{\textit{Pass}}_0$, etc. 
If $\textbf{\textit{Pass}}_0=\textbf{\textit{False}}$, the research agent will generate the
error correction strategy.
%
%
Then the code learner agent and testbench learner agent will correct the codes with strategies generated from the research agent.
Whereafter, Iverilog simulation $\textbf{\textit{I}}_v$ is utilized to validate the effectiveness of the modifications through evaluating the correctness of both code $\textbf{\textit{Code}}_{k+1}$ and testbench $\textbf{\textit{TB}}_{k+1}$ modifications in the \textit{\textbf{k}}th iteration of error correction, which can be denoted as:
\begin{align*}
   \textbf{\textit{Pass}}_{k+1},\textbf{\textit{E}}_{k+1} = \textbf{\textit{I}}_v(\textbf{\textit{Code}}_{k+1},\textbf{\textit{TB}}_{k+1})
\end{align*}

$\textbf{\textit{Pass}}_{k+1}$ and error information $\textbf{\textit{E}}_{k+1}$ can be obtained after the Iverilog simulation.
If $\textbf{\textit{Pass}}_{k+1}$ is  \textbf{\textit{False}}, 
the next round of correction for Verilog code and testbench begins.
The erroneous code and testbench are returned to the research agent, which performs the next round of corrections.

\begin{table*}[!ht]
\caption{Comparison of CoopetitiveV against various baseline models }
\vspace{-0.1in}
\centering
\resizebox{\textwidth}{!}{  
\begin{tabular}{|c|c|c|cccccc|cc|}
\toprule
\multirow{3}{*}{Type}                & \multirow{3}{*}{Model} & \multirow{3}{*}{\begin{tabular}[c]{@{}c@{}}Open \\ source\end{tabular}} & \multicolumn{6}{c|}{VerilogEval}                                                                                                                               & \multicolumn{2}{c|}{RTLLM pass@5}                                            \\ \cline{4-11} 
                                     &                        &                                                                         & \multicolumn{3}{c|}{Machine(\%)}                                                         & \multicolumn{3}{c|}{Human(\%)}                                      & \multicolumn{1}{c|}{\multirow{2}{*}{Syntax(\%)}} & \multirow{2}{*}{Func(\%)} \\ \cline{4-9}
                                     &                        &                                                                         & \multicolumn{1}{c|}{pass@1} & \multicolumn{1}{c|}{pass@5} & \multicolumn{1}{c|}{pass@10} & \multicolumn{1}{c|}{pass@1} & \multicolumn{1}{c|}{pass@5} & pass@10 & \multicolumn{1}{c|}{}                            &                           \\ \midrule
\multirow{6}{*}{Foundation models}   & GPT-3.5                & N                                                                       & \multicolumn{1}{c|}{46.7}   & \multicolumn{1}{c|}{69.1}   & \multicolumn{1}{c|}{74.1}    & \multicolumn{1}{c|}{26.7}   & \multicolumn{1}{c|}{45.8}   & 51.7    & \multicolumn{1}{c|}{89.7}                        & 37.9                      \\ \cline{2-11} 
                                     & GPT-4                  & N                                                                       & \multicolumn{1}{c|}{60.0}   & \multicolumn{1}{c|}{70.6}   & \multicolumn{1}{c|}{73.5}    & \multicolumn{1}{c|}{43.5}   & \multicolumn{1}{c|}{55.8}   & 58.9    & \multicolumn{1}{c|}{100}                         & 65.5                      \\ \cline{2-11} 
                                     & Claude-3               & N                                                                       & \multicolumn{1}{c|}{55.3}   & \multicolumn{1}{c|}{63.8}   & \multicolumn{1}{c|}{69.4}    & \multicolumn{1}{c|}{34.4}   & \multicolumn{1}{c|}{48.3}   & 53.4    & \multicolumn{1}{c|}{93.1}                        & 55.2                      \\ \cline{2-11} 
                                     & CodeLlama              & Y                                                                       & \multicolumn{1}{c|}{43.1}   & \multicolumn{1}{c|}{47.1}   & \multicolumn{1}{c|}{47.7}    & \multicolumn{1}{c|}{18.2}   & \multicolumn{1}{c|}{22.7}   & 24.3    & \multicolumn{1}{c|}{86.2}                        & 31.0                      \\ \cline{2-11} 
                                     & DeepSeek-Coder         & Y                                                                       & \multicolumn{1}{c|}{52.2}   & \multicolumn{1}{c|}{55.4}   & \multicolumn{1}{c|}{56.8}    & \multicolumn{1}{c|}{30.2}   & \multicolumn{1}{c|}{33.9}   & 34.9    & \multicolumn{1}{c|}{93.1}                        & 44.8                      \\ \cline{2-11} 
                                     & CodeQwen               & Y                                                                       & \multicolumn{1}{c|}{46.5}   & \multicolumn{1}{c|}{54.9}   & \multicolumn{1}{c|}{56.4}    & \multicolumn{1}{c|}{22.5}   & \multicolumn{1}{c|}{26.1}   & 28.0    & \multicolumn{1}{c|}{86.2}                        & 41.4                      \\ \midrule
\multirow{10}{*}{Specialized models} & ChipNeMo-13B           & N                                                                       & \multicolumn{1}{c|}{43.4}   & \multicolumn{1}{c|}{-}      & \multicolumn{1}{c|}{-}       & \multicolumn{1}{c|}{22.4}   & \multicolumn{1}{c|}{-}      & -       & \multicolumn{1}{c|}{-}                           & -                         \\ \cline{2-11} 
                                     & ChipNeMo-70B           & N                                                                       & \multicolumn{1}{c|}{53.8}   & \multicolumn{1}{c|}{-}      & \multicolumn{1}{c|}{-}       & \multicolumn{1}{c|}{27.6}   & \multicolumn{1}{c|}{-}      & -       & \multicolumn{1}{c|}{-}                           & -                         \\ \cline{2-11} 
                                     & RTLCoder-Mistral       & Y                                                                       & \multicolumn{1}{c|}{62.5}   & \multicolumn{1}{c|}{72.2}   & \multicolumn{1}{c|}{76.6}    & \multicolumn{1}{c|}{36.7}   & \multicolumn{1}{c|}{45.5}   & 49.2    & \multicolumn{1}{c|}{96.6}                        & 48.3                      \\ \cline{2-11} 
                                     & RTLCoder-DeepSeek      & Y                                                                       & \multicolumn{1}{c|}{61.2}   & \multicolumn{1}{c|}{76.5}   & \multicolumn{1}{c|}{81.8}    & \multicolumn{1}{c|}{41.6}   & \multicolumn{1}{c|}{50.1}   & 53.4    & \multicolumn{1}{c|}{93.1}                        & 48.3                      \\ \cline{2-11} 
                                     & BetterV-CodeLlama      & N                                                                       & \multicolumn{1}{c|}{64.2}   & \multicolumn{1}{c|}{75.4}   & \multicolumn{1}{c|}{79.1}    & \multicolumn{1}{c|}{40.9}   & \multicolumn{1}{c|}{50.0}   & 53.3    & \multicolumn{1}{c|}{-}                           & -                         \\ \cline{2-11} 
                                     & BetterV-DeepSeek       & N                                                                       & \multicolumn{1}{c|}{67.8}   & \multicolumn{1}{c|}{79.1}   & \multicolumn{1}{c|}{84.0}    & \multicolumn{1}{c|}{45.9}   & \multicolumn{1}{c|}{53.3}   & 57.6    & \multicolumn{1}{c|}{-}                           & -                         \\ \cline{2-11} 
                                     & BetterV-CodeQwen       & N                                                                       & \multicolumn{1}{c|}{68.1}   & \multicolumn{1}{c|}{79.4}   & \multicolumn{1}{c|}{84.5}    & \multicolumn{1}{c|}{46.1}   & \multicolumn{1}{c|}{53.7}   & 58.2    & \multicolumn{1}{c|}{-}                           & -                         \\ \cline{2-11} 
                                     & CodeV-CodeLlama        & Y                                                                       & \multicolumn{1}{c|}{78.1}   & \multicolumn{1}{c|}{86.0}   & \multicolumn{1}{c|}{88.5}    & \multicolumn{1}{c|}{45.2}   & \multicolumn{1}{c|}{59.5}   & 63.8    & \multicolumn{1}{c|}{93.1}                        & 62.1                      \\ \cline{2-11} 
                                     & CodeV-DeepSeek         & Y                                                                       & \multicolumn{1}{c|}{77.9}   & \multicolumn{1}{c|}{88.6}   & \multicolumn{1}{c|}{90.7}    & \multicolumn{1}{c|}{52.7}   & \multicolumn{1}{c|}{62.5}   & 67.3    & \multicolumn{1}{c|}{89.7}                        & 55.2                      \\ \cline{2-11} 
                                     & CodeV-CodeQwen         & Y                                                                       & \multicolumn{1}{c|}{77.6}   & \multicolumn{1}{c|}{88.2}   & \multicolumn{1}{c|}{90.7}    & \multicolumn{1}{c|}{53.2}   & \multicolumn{1}{c|}{65.1}   & 68.5    & \multicolumn{1}{c|}{93.1}                        & 55.2      \\
                                     
                                  \midrule
\multirow{2}{*}{CoopetitiveV (Ours)}          & Ours + GPT-3.5         & Y                                                                       & \multicolumn{1}{c|}{56.3}   & \multicolumn{1}{c|}{92.1}   & \multicolumn{1}{c|}{96.7}    & \multicolumn{1}{c|}{49.2}   & \multicolumn{1}{c|}{88.3}   & 96.4    & \multicolumn{1}{c|}{99.8}                        & 88.8                      \\ \cline{2-11} 
                                     & Ours + GPT-4           & Y                                                                       & \multicolumn{1}{c|}{\textbf{93.9}}   & \multicolumn{1}{c|}{\textbf{99.0}}   & \multicolumn{1}{c|}{\textbf{99.2}}    & \multicolumn{1}{c|}{\textbf{89.3}}   & \multicolumn{1}{c|}{\textbf{98.5}}   & \textbf{99.1}    & \multicolumn{1}{c|}{\textbf{100}}                         & \textbf{99.9}      \\ \cline{2-11}
                                     
                                     & Ours + Claude-3.5           & Y                                                                       & \multicolumn{1}{c|}{\textbf{97.8}}   & \multicolumn{1}{c|}{\textbf{99.2}}   & \multicolumn{1}{c|}{\textbf{99.3}}    & \multicolumn{1}{c|}{\textbf{94.9}}   & \multicolumn{1}{c|}{\textbf{97.9}}   & \textbf{99.0}    & \multicolumn{1}{c|}{\textbf{100}}                         & \textbf{99.8}  
                                     \\ \bottomrule 
\end{tabular}
}
\label{table:main}
\end{table*}

\begin{table}[ht]
\centering
\caption{Results of pass@1 on Verilog-Human and Verilog-V2 datasets using Claude 3.5 model}
\vspace{-0.1in}
\label{tab:verilog_pass_rates}
\renewcommand{\arraystretch}{1.3}
\resizebox{\linewidth}{!}{%
\begin{tabular}{|l|l|c|c|}
\hline
\makecell{\textbf{Method}} & \textbf{LLM Model} & \makecell{\textbf{Verilog-Human}\\\textbf{Pass@1}} & \makecell{\textbf{Verilog-V2}\\\textbf{Pass@1}} \\
\hline
Generic LLM     & Claude 3.5  & 75.0 & 72.4 \\
\hline
Mage            & Claude 3.5 & 94.8 & 95.7 \\
\hline
CoopetitiveV (Ours)    & Claude 3.5  & \textbf{94.9} & \textbf{96.0} \\
\hline
\end{tabular}

} 

\vspace{0.5em}
\footnotesize{
*We use Claude-3.5 Sonnet 2024-10-22 as our base model
}
\end{table}

\section{Experiments}

\subsection{Experimental Setup}

\subsubsection{Datasets and Baselines}


\textbf{Datasets: }We evaluate CoopetitiveV on
three comprehensive datasets: VerilogEval~\cite{liu2023verilogeval}, VerilogEval-V2~\cite{ho2025verilogcoder} and RTLLM~\cite{lu2024rtllm}.
The VerilogEval comprises VerilogEval-machine and VerilogEval-human. 
\textbf{Baselines: }We evaluate our method against two types of baseline models,  including: (1) general-purpose foundation models such as GPT-3.5, GPT-4, and Claude-3, alongside three open-source models designed for code generation, namely CodeLlama-7B-Instruct\cite{roziere2023code}, DeepSeek-Coder-6.7B-Instruct \cite{guo2024deepseek}, and CodeQwen-1.5-7B-Chat \cite{bai2023qwen}; (2) specifically tuned models for hardware code generation, such as ChipNeMo \cite{liu2023chipnemo}, RTLCoder \cite{liu2024rtlcoder}, BetterV \cite{pei2024betterv}, and CodeV\cite{zhao2024codev}. We also compare our method with Mage~\cite{zhao2024mage}, one of the SOTA LLM-assisted multi-agent learning method for Verilog generation.

\subsubsection{Metrics}

We follow VerilogEval~\cite{liu2023verilogeval} and adopt pass@k as the evaluation metric on various datasets. The pass@k metric is expressed as 
\begin{equation}
   \operatorname{pass} @ k:=\underset{\text { problems }}{\mathbb{E}}\left[\frac{1-\binom{n-c}{k}}{\binom{n}{k}}\right] 
\end{equation}

where $n$ denotes the total number of trials for each problem, $c$ refers to the number of passed trials for a given problem and $k$ denotes the number of trials selected to be evaluated. The pass@\textit{k} metric calculates the proportion of problems that can be solved at least once in \textit{k} samples. In our experiments, we set \( n = 20 \). 

\subsection{Experimental Results}
The main results of CoopetitiveV are shown in Table~\ref{table:main}. We compare it with baselines across the VerilogEval (including VerilogEval-Machine and VerilogEval-Human) and RTLLM benchmarks. On VerilogEval, we use pass@1, pass@5, and pass@10 as the evaluation metrics,
while on RTLLM, we adopt pass@5,
same with existing works~\cite{zhao2024codev, Huang2024TowardsLV}. Our approach achieves SOTA on both VerilogEval and RTLLM benchmarks. 
On VerilogEval-Machine, CoopetitiveV+GPT-3.5 has 9.6\%, 23.0\%, and 22.6\% higher scores than GPT-3.5 on pass@1, pass@5, pass@10, respectively.
CoopetitiveV+GPT-4 scores 16.0\%, 10.4\%, and 8.5\% higher than the SOTA on pass@1, pass@5, and pass@10, respectively.
On VerilogEval-Human,
CoopetitiveV+GPT-3.5 has 22.5\%, 42.5\%, and 44.7\% higher scores than GPT-3.5 on pass@1, pass@5, pass@10, respectively. 
CoopetitiveV+GPT-4 scores 45.8\%, 42.7\%, and 40.2\% higher than GPT-4, while obtaining 36.0\% and 31.8\% higher scores compared to
all baselines on pass@5, and pass@10, respectively.
%
On RTLLM pass@5, CoopetitiveV+GPT-3.5 shows an advantage over the GPT-3.5, with 10.1\% and 51.1\% higher scores for syntax and functionality checking.
CoopetitiveV+GPT-4 and CoopetitiveV+Claude-3.5 wins all the baselines, both with around 100\% on syntax and functionality checking on pass@5. In Table \ref{tab:verilog_pass_rates}, we compare our framework with Mage on Verilog-Human and Verilog-V2, our pipeline demonstrates a slight advantage over MAGE, achieving an improvement of approximately 0.3 percentage points.

We observe that CoopetitiveV has great advantage on pass@5 and pass@10 for both VerilogEval-Machine and VerilogEval-Human. We consider the reason as follows: 
our integration of multi-agent learning, specifically the teacher agent component, provides targeted guidance for Verilog code correction. This approach ensures a higher pass rate on individual code samples, substantially reducing the probability of complete failure across all 20 trials. 
While our method may not guarantee very high pass probability for each sample within individual trials, it consistently achieves at least several successful passes across the 20 trails for each dataset entry. This characteristic explains the observed pattern in our results: relatively lower pass@1 rates on the VerilogEval datasets, but significantly higher pass@5 and pass@10 rates, demonstrating the robustness of our approach over multiple attempts.

Besides the advantages on pass@5 and pass@10, CoopetitiveV+GPT-4 and CoopetitiveV+Claude-3.5 also demonstrates 
relatively high scores on pass@1, while significantly reducing the validation gap between VerilogEval-Machine and VerilogEval-Human. 
We consider two key factors driving these improvements:
First, both CoopetitiveV+GPT-4 and CoopetitiveV+Claude-3.5 exhibit enhanced accuracy in initial code generation, further refined through our iterative correction process, resulting in more precise Verilog code. 
Second, GPT-4 and Claude-3.5's advanced language processing capabilities and expanded token context windows enable comprehensive processing of complex inputs, particularly Autobench-generated long token sequence stimulus signals that exceed GPT-3.5's capacity limits.

\subsection{Ablation Study}

\begin{figure}
    \centering
   \includegraphics[width=0.99\columnwidth]{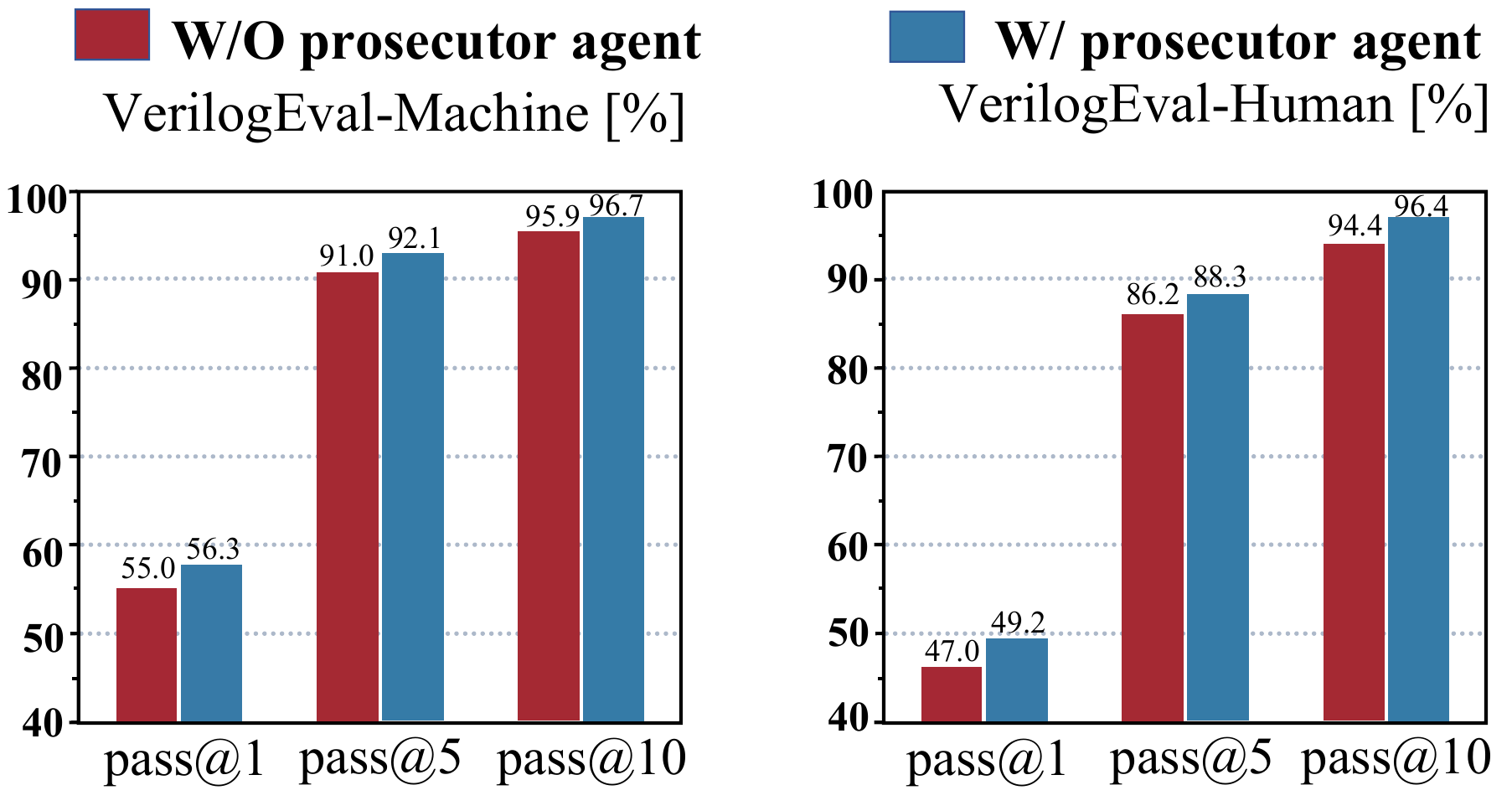}
    \vspace{-0.3in}
    \caption{Comparison of without and with prosecutor agent with CoopetitiveV+GPT3.5: (a) On VerilogEval-Machine, (b) On VerilogEval-Human}
    \vspace{-0.2in}
    \label{fig:ablation_teacher}
\end{figure}

\textbf{The effects of competition mechanism.} 
To explore the effects of the competition mechanism on Verilog code generation and correction, we design the following experiments
with GPT-3.5 on VerilogEval-Machine and VerilogEval-Human: 
1) w/o prosecutor agent: we exclude the prosecutor agent from the framework and rely solely on the cooperation mechanism, which involves only the two revision agents and the research agents for Verilog code generation and correction. 
2) w/ prosecutor agent: the prosecutor agent challenges the correction strategy generated
by the research agent, and generate the improved code revision strategies whenever possible. 
The comparison experiments aim to evaluate the effects of the prosecutor agent in identifying and correcting code errors.
The results are shown in Figure~\ref{fig:ablation_teacher}. 
It's shown that with the integration of prosecutor agent, we can achieve higher pass@1, pass@5, and pass@10 scores on 
both VerilogEval-machine and VerilogEval-human datasets. The improved accuracy 
indicates that the competition mechanism can effectively enhance the Verilog quality generated by the revision agents.

\section{Related work}

\textbf{LLM-based Verilog Generation.}
Recent advancements in LLMs have demonstrated their significant potential in code generation tasks.
While early research primarily focuses on generating functional programming code, exemplified by OpenAI’s Codex~\cite{Chen2021EvaluatingLL}, the application of LLMs to hardware design, particularly Verilog generation, is gaining increasing attention. 
~\cite{liu2023chipnemo} introduces domain-adapted LLMs for chip design, enhancing the capability of LLMs in handling hardware-related tasks. 
~\cite{liu2024rtlcoder} utilizes a lightweight transformer model optimized with a custom open-source RTL dataset, significantly improving efficiency and accuracy in RTL design generation compared to GPT-3.5. However, these works suffer from 
limited generation quality.

\textbf{LLM-based Agentic Learning.}
Efforts have been developed on LLM-based code generation automation~\cite{zan2022large,pearce2020dave}.
Initial approaches utilize a single LLM agent for code generation and correction~\cite{wang2024survey, shen2023pangu, madaan2024self, wang2022self}.
However, 
single-agent 
have constrained capability in dealing with code-related tasks, especially
in code generation~\cite{madaan2024self}.
The limitation of single-agent learning motivates the exploration
of multi-agent learning~\cite{huang2023agentcoder}
across various tasks, including code generation~\cite{shrivastava2023repository}, code error correction~\cite{dearing2024lassi, ugare2024improving}, code logic optimization~\cite{zhang2023self, zhang2023coder, gao2024search}, etc.
For example, ~\cite{wang2024intervenor} employs LLM-based agents with distinct roles of teacher and learner in code repairing. 
~\cite{huang2023codecot} proposes Code Chain-of-Thought (CodeCoT) that integrates CoT with a self-examination process for code generation.

\vspace{-0.4em}
\section{Conclusion}
We propose CoopetitiveV, an interactive LLM-powered multi-agent framework for high-quality Verilog generation. CoopetitiveV integrates the proposed cooperation and competition mechanism to better operate multiple specialized LLM agents, each of which dedicates to distinct tasks, including code generation, strategy proposing, strategy challenging and code error correction, effectively mitigating the degeneration issue and error propagation in traditional LLM-based pipeline. 
Our comprehensive empirical evaluation across various benchmarks demonstrates the effectiveness of the proposed method in Verilog generation, sheding light on the automation of EDA tasks with the power of LLMs. 

\section*{Limitations}

Our method currently focuses solely on Verilog-based code generation task, though the coopetitive mechanism could potentially extend to other domains such as code-to-code translation and cross-language synthesis. In future work, we aim to explore its applicability in a wider range of areas.



\bibliography{custom}




\end{document}